\documentclass[journal]{IEEEtran}

\usepackage{algorithm}
\usepackage{algorithmicx}
\usepackage[noend]{algpseudocode} 

\usepackage{cite}
\usepackage{acro}
\usepackage{graphicx}
\usepackage{float}
\usepackage{xcolor}
\usepackage{amsmath}

\newcommand{\Fig}{Fig.}

\newcommand{\Tab}{Table}
\newcommand{\etal}{\textit{et al.}}

\DeclareAcronym{FCN}{
short=FCN,
long=fully convolutional network,
}

\DeclareAcronym{CNN}{
short=CNN,
long=convolutional neural network,
}

\DeclareAcronym{MOT}{
short=MOT,
long=multiple object tracking,
}

\DeclareAcronym{SOT}{
short=SOT,
long=single object tracking,
}

\DeclareAcronym{LSTM}{
short=LSTM,
long=long short-term memory,
}

\DeclareAcronym{DPA}{
short=DPA,
long=dual-path association,
}

\DeclareAcronym{Map3D}{
short=Map3D,
long=Multi-object Association for Pathology in 3D,
}

\DeclareAcronym{IoU}{
short=IoU,
long=Intersection Over Union,
}

\DeclareAcronym{SIFT}{
short=SIFT,
long=scale-invariant feature transform,
}

\DeclareAcronym{SG}{
short=SG,
long=SuperGlue,
}

\DeclareAcronym{QaWS}{
short= QaWS,
long=quality-aware whole series ,
}

\DeclareAcronym{QA}{
short= QA,
long=quality assurance,
}

\DeclareAcronym{ANTs}{
short= ANTs,
long=advanced normalization tools,
}

\DeclareMathOperator*{\argminA}{arg\,min} %

\begin{document}
%
\title{Map3D: Registration Based Multi-Object Tracking on 3D Serial Whole Slide Images}
%
%
%

\author{Ruining~Deng,
        Haichun~Yang,
        Aadarsh~Jha,
        Yuzhe~Lu,
        Peng~Chu,
        Agnes~B.~Fogo,
        and~Yuankai~Huo*,~\IEEEmembership{Member,~IEEE}
\thanks{*Y. Huo is the corresponding author, e-mail: yuankai.huo@vanderbilt.edu}
\thanks{R. Deng, A. Jha, Y. Lu,  Y. Huo were with the Department of Computer Science, Vanderbilt University, Nashville, TN, 37215, USA, }
\thanks{H. Yang, A. Fogo were with the Department
of Pathology, Vanderbilt University Medical Center, Nashville,
TN, 37215, USA}
\thanks{P. Chu was with the Department
of Computer and Information Sciences, Temple University,Philadelphia, PA 19122}
}

\markboth{Manuscript pre-print, June~2020}%
{Shell \MakeLowercase{\textit{et al.}}: Bare Demo of IEEEtran.cls for IEEE Journals}

\maketitle

\begin{abstract}
There has been a long pursuit for precise and reproducible glomerular quantification on renal pathology to leverage both research and practice. When digitizing the biopsy tissue samples using whole slide imaging (WSI), a set of serial sections from the same tissue can be acquired as a stack of images, similar to frames in a video. In radiology, the stack of images (e.g., computed tomography) are naturally used to provide 3D context for organs, tissues, and tumors. In pathology, it is appealing to do a similar 3D assessment. However, the 3D identification and association of large-scale glomeruli on renal pathology is challenging due to large tissue deformation, missing tissues, and artifacts from WSI. In this paper, we propose a novel Multi-object Association for Pathology in 3D (Map3D) method for automatically identifying and associating large-scale cross-sections of 3D objects from routine serial sectioning and WSI. The innovations of the Map3D method are three-fold: (1) the large-scale glomerular association is formed as a new multi-object tracking (MOT) perspective; (2) the quality-aware whole series registration is proposed to not only provide affinity estimation but also offer automatic kidney-wise quality assurance (QA) for registration; (3) a dual-path association method is proposed to tackle the large deformation, missing tissues, and artifacts during tracking. To the best of our knowledge, the Map3D method is the first approach that enables automatic and large-scale glomerular association across 3D serial sectioning using WSI. Our proposed method Map3D achieved MOTA= 44.6, which is 12.1$\%$ higher than the non deep learning benchmarks.
\end{abstract}


%
\IEEEpeerreviewmaketitle

\section{Introduction}
\IEEEPARstart{O}{ver} the past decade, rapid advances in whole slide imaging (WSI) and image processing have led to a paradigm shift in analyzing large-scale high-resolution renal pathology images~\cite{rangan2007quantification}. These advances are largely attributed to progress in deep learning techniques, which have enabled high throughput object quantification for clinical research and practice. However, current quantitative assessments of glomeruli are still primarily performed on a single two-dimensional (2D) section, which is error-prone due to the heterogeneity of glomeruli across serial sections. For example, a recent study~\cite{fogo1995focal} elucidated that 2D phenotyping of the percentage of glomerulosclerosis could be misleading compared with 3D phenotyping. Moreover, several important glomerular phenotypes are ideally, or of necessity, gained in 3D, such as glomerular volume and atubular glomeruli, respectively. 

\begin{figure*}[t]
\begin{center}
\includegraphics[width=0.9\textwidth]{{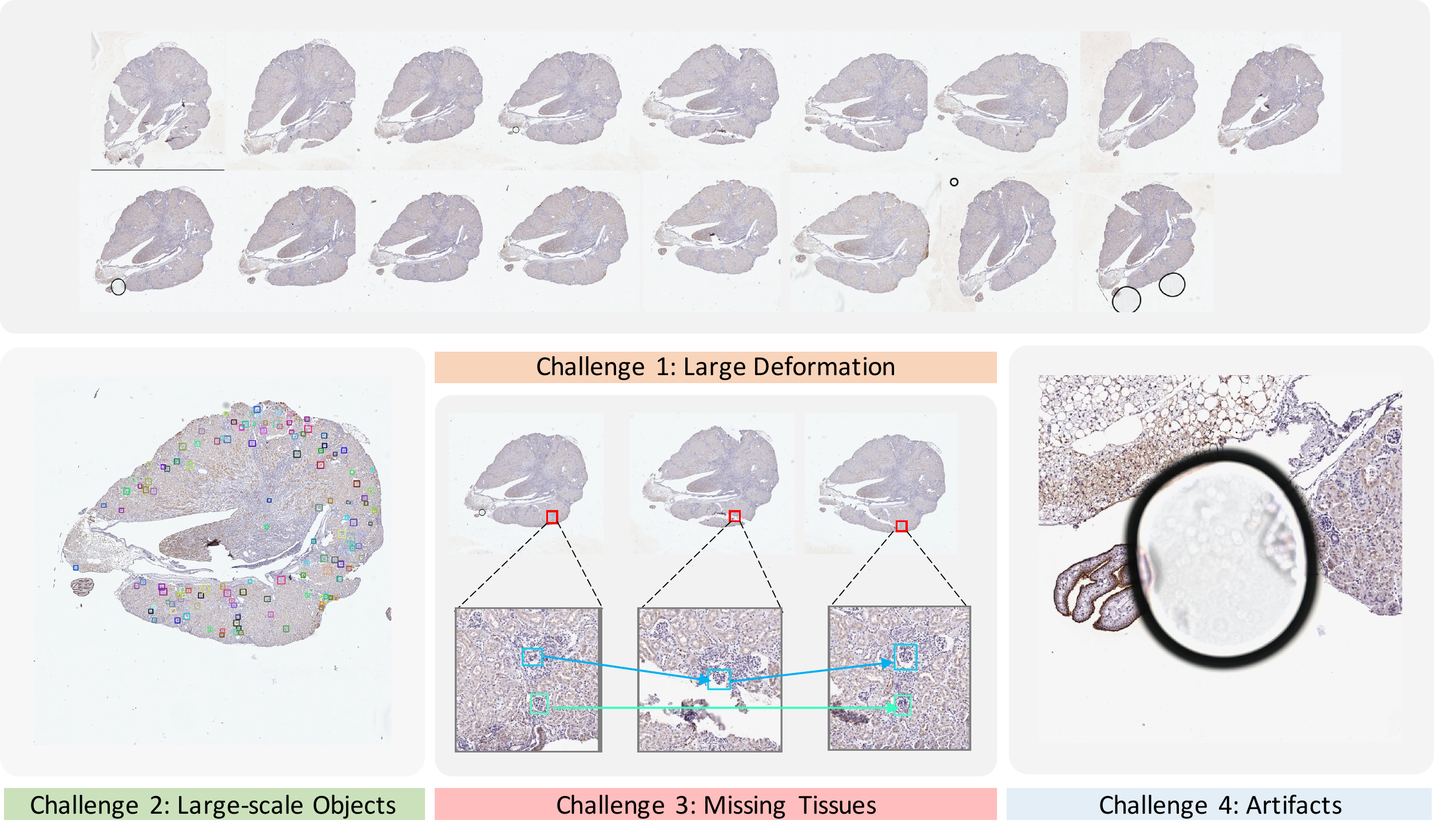}}
\end{center}
\caption{This figure shows the challenges of performing multi-object tracking (\ac{MOT}) for large-scale glomerular identification and association on serial sectioned whole slide images (WSI).} 
\label{Fig.1} 
\end{figure*}

Atubular glomeruli are glomeruli that have lost connection with the proximal tubule, severely decreasing the glomerular filtration rate and affecting kidney disease progression. In animal model (i.e, mice model), this pathological changes can only be confirmed when all WSI sections of a nephron are visually examined by 3D~\cite{chevalier2008generation} to identify atubular glomerulus by tracking individual glomeruli in serial sections. Even though the 3D assessments are more precise and reproducible, it is technically challenging to perform scalable 3D glomerular quantification on kidney WSI, since a considerable number of 2D glomerular cross-sections (image patches) need to be associated from serial sectioning, along with large tissue deformation, tissue missing, and artifacts from tissue sectioning and imaging ({\color{red} \Fig~\ref{Fig.1}}). As a result, current 3D glomerular studies are still relying heavily on manual or semi-automated approaches, leading to increased labor costs, low-throughput image analysis and potential inter-observer variability. 

The robust biological image registration is significant to the 3D identification and association of glomeruli, providing more precise data to pathologists and clinicians. Wang, Ching-Wei et al ~\cite{wang2014robust} have proposed an automatic and fast image registration method to alleviate the registration error caused by artifacts, tissue splitting, and tissue folding problems in WSI images, while achieving robust registration results. Ali S et al. ~\cite{ali2018rigid} have introduced a novel multi-modal similarity metric and an improved regularization scheme to tackle deformations. SURF ~\cite{bay2006surf} was employed in this study to improve initial alignment. Cooper et al., ~\cite{cooper2009feature} have developed a method to use the automatic non-rigid registration on histological section images with different stains to tackle the intensity inconsistency issue. These methods have improved the registration results. However, such methods have not dealt with object tracking across serial sections, with imaging artifacts and missing tissue.

Here, we principle a fully automatic large-scale 3D glomerular identification and association from a new \ac{MOT} perspective, splitting the challenging task to consequential steps (object detection, affinity estimation, and 3D association). However, there are still unique challenges for developing \ac{MOT} on renal pathology as opposed to the canonical \ac{MOT} tasks into computer vision. For example, the resolution of a pathology image is in orders of magnitude higher than typical natural images, bringing challenges in detection and association. Large deformation, missing tissues, and artifacts (Fig. 1) are typically inevitable during section preparation and imaging to cause bad registration performances with on evaluation processes. Moreover, no large-scale annotated training data are publicly available, impeding the utilization of deep learning based \ac{MOT} algorithms. To address these challenges, a pretrained Graph Neural Network based SuperGlue keypoints matching and registration based association are aggregated to offer high throughput object detection and annotation free association. The proposed method is enabled by taking the advantages from both computer vision and medical image processing, without using any domain specific training data.

In this paper, we propose a \ac{Map3D} method, for the scalable and automatic glomerular association in 3D renal pathology. Briefly, our previously proposed CircleNet method~\cite{yang2020circle} is employed for large-scale glomerular detection. Then, a registration-based affinity estimation method, called \ac{QaWS} registration, is developed to not only estimate the pixel-wise affinity across the different sections, but also offer automatic \ac{QA} for the entire stack of images. The automatic \ac{QA} is important when deploying the \ac{Map3D} on a large-scale dataset. Last, the \ac{DPA} algorithm is introduced to associate all detected glomerular cross-sections in a 3D context, addressing the continuous tracking for missing tissues and artifacts. Serial whole kidney sections from 14 mice were used to train and validate the performance of the proposed method. These experiments show that \ac{Map3D} is a promising step towards the ultimate goal of reducing labor costs and time needed involved in densely annotating and associating all glomeruli from serial sections in one kidney from a manual (30 hours per kidney) to a fully automatic manner.

To summarize, the innovations of the \ac{Map3D} method are three-fold: (1) a novel holistic \ac{MOT} framework is proposed to address the challenging 3D glomerular identification and association on high-resolution images with large deformation; (2) the \ac{QaWS} registration, with Graph Neural Network based SuperGlue keypoints matching, is proposed to not only provide affinity estimation but also offer automatic kidney-wise quality assurance (\ac{QA}) for registration; (3) a Dual-path 3D association method \ac{DPA} is presented to tackle the presence of missing tissues and artifacts during serial sectioning that are then scanned as WSI. To the best of our knowledge, the \ac{Map3D} method is the first method that works toward automatic and large-scale glomerular identification and 3D associations across routine serial sectioning WSI. 

\begin{figure*}[t]
\centering 
\includegraphics[width=0.9\textwidth]{{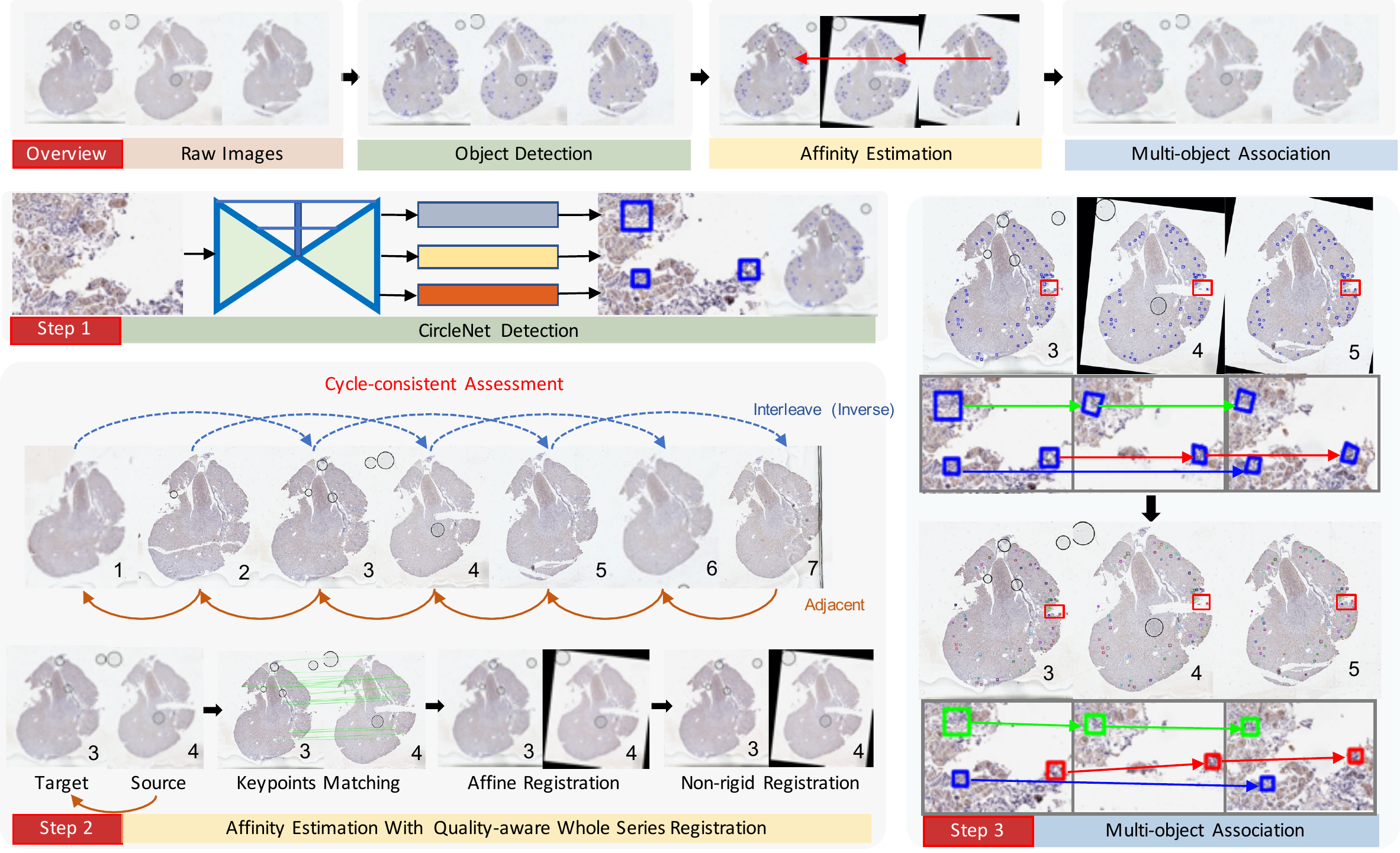}}
\caption{The overview as well as each step are presented in this figure. The overview panel shows the three major steps in our MOT framework: (1) object detection, (2) affinity estimation, and (3) multi-object association. In step 1 object detection, deep learning based high-throughput detection method is used to detect all glomeruli. In step 2 affinity estimation, both affine and non-rigid registration are used to achieve pixel-wise correspondence between sections. In step 3 multi-object association, the dual-path association (\ac{DPA}) is used to perform object tracking with missing tissues.} 
\label{Fig.2} 
\end{figure*}

\section{Related Works}
\subsection{Multi-object Tracking}
\ac{MOT} has been an essential research area in computer vision for decades. Its primary aim is to track multiple objects from a video. Besides tackling scale, rotation, and intensity issues in Single-object Tracking (SOT), \ac{MOT} focuses more on identifications and interactions among multiple similar objects. Furthermore, the advances of deep learning have changed the paradigm of \ac{MOT} from a model based policy to data-driven approaches~\cite{xiang2015data}. The current focus has been centered on a ``tracking-by-detection" principle. The \ac{MOT} algorithms can be roughly classified into two families. 

The first family treats \ac{MOT} as an online estimation study since real time performance is required in many computer vision tasks, such as self-driving, video surveillance, and cell phone applications. In such scenarios, the tracking status of the current frame is determined by previous observations until the current time point as an online learning procedure~\cite{yan2012track,xiang2015learning,sadeghian2017tracking}. Yan \etal~\cite{yan2012track} captured target candidates from both detector and independent single object trackers, by integrating the messages to determine optimal tracking. Xiang \etal~\cite{xiang2015learning} deployed the \ac{MOT} as a Markov decision process using annotated training data. 

The second family tackles \ac{MOT} as a global optimization problem using offline optimization by utilizing both previous and future slides to determine the current status of tracking. The most commonly used global data association algorithms are the Hungarian algorithm~\cite{bewley2016simple,fang2018recurrent}, multiple hypotheses tracking~\cite{kim2015multiple}, and network flow~\cite{zamir2012gmcp,dehghan2015target}. The quality of the tracking is largely reliant on the accuracy of detected ions from the external detector.

Recently, deep learning has been widely used in \ac{MOT} due to its high accuracy and computational efficiency. Most recent solutions rely on a powerful discriminant technique~\cite{son2017multi,zhu2018online} for robust affinity estimation. Tang \etal~\cite{tang2017multiple} proposed a deep learning based affinity estimation method. Sadeghian \etal~\cite{sadeghian2017tracking} employed a \ac{CNN} and a \ac{LSTM} to model long-term temporal dependencies by aggregating clues from interaction, motion, and a person re-identification model using a dynamic \ac{CNN}-based framework. Recently, end-to-end deep learning solutions have been used ~\cite{chu2019online,xu2019deepmot} to improve the performance.

In this paper, we define object association in 3D renal pathology as an \ac{MOT} task. In contrast to latest \ac{MOT} studies, no large-scale training data are available to train a deep learning based solution. Moreover, the sections in the pathology ``video" (serial sections) have global rotation, deformation, and artifacts. Therefore, we propose the registration based \ac{MOT} method, inspired by the recent innovations in 3D registration and reconstruction on pathological images~\cite{rossetti2017dynamic,pichat2015multi,liang2017development}. Unlike these studies, which achieve the ``perfect" 3D reconstruction of the entire WSI stack, we employ the registration as an intermediate step to estimate the affinity between glomerular detection results. Therefore, we only emphasize the registration across neighboring sections (frames) in an \ac{MOT} context and eventually assemble pair-wise alliances to global identifications and associations. 


\subsection{Deep Learning Based Glomerular Quantification}
WSI represents a paradigm shift, enabling clinicians to diagnose patients and guide therapeutic planning by navigating a virtual slide. Imaging advances have driven increasing demands in high throughput image quantification for clinical decision support. Excitingly, the explosive growth in deep learning technologies has been adapted to the field of renal pathology to match such needs~\cite{sarwar2019physician}. Many deep learning studies have been focused on glomerular quantification since this parameter and the study of glomerular lesions are essential in nephrology. 
The current glomerular quantification methods are mostly 2D-based quantifications, where tasks can be categorized as classification~\cite{ginley2019computational,uchino2020classification}, detection~\cite{simon2018multi,maree2016approach}, and segmentation~\cite{bueno2020glomerulosclerosis,kannan2019segmentation,gadermayr2017segmenting,ginley2019computational}. Beyond basic quantification, many recent works have performed further analyses based upon such preliminary quantifications~\cite{marsh2018deep,barros2017pathospotter}. A recent study even provided the dense estimation of a renal pathology image including a comprehensive notation of ten tissue classes ~\cite{hermsen2019deep}. Distinct from previous 2D glomerular quantification methods, we propose a 3D quantification framework, combining the deep learning and \ac{MOT}.

\section{Methods}
The entire framework of the proposed \ac{Map3D} is presented in {\color{red} \Fig~\ref{Fig.2}}. The \ac{Map3D} pipeline consists of three sections: (1) glomerular detection, (2) \ac{QaWS} registration based affinity estimation, and (3) \ac{DPA} based 3D association.


\subsection{Object Detection}
Glomerular detection was implemented by our previously proposed CircleNet method~\cite{yang2020circle}, which has the aim to develop optimized bounding circle representations for glomerular detection. As the CircleNet achieved superior performance for glomerular detection compared with current benchmarks, we directly applied CircleNet as the detection method in this study. Due to the scarcity dataset, we trained our original CircleNet model human glomeruli [8], whose structure is similar to the mice. To adapt the detection method to mouse kidneys, we fine-tuned the CircleNet using image patches from 927 and 125 mice glomeruli as training and validation data (independent from the validation and testing data in the \ac{MOT} task). As an \ac{MOT} design, all the detection results are saved as bounding boxes with their corner coordinates. The similarities between different bounding boxes are measured by \ac{IoU}.

\subsection{Quality-aware Whole Series Registration}
After achieving the bounding boxes from detection, the standard operation in \ac{MOT} is to calculate affinity measurements across detected objects. In the pathological WSI image, if the 3D serial sectioned images are regarded as a video, the unique challenges are the high resolution of each video ``frame" (i.e., one section) and the deformation between cross-sections in different depth from one glomerulus. However, there are also unique benefits for tracking objects across sections in WSI. For example, the relative locations of different tissues are more stable than the \ac{MOT} tasks in computer vision. Inspired by these facts and a previous study, we decided to used image registration as the affinity estimation method~\cite{rossetti2017dynamic}. 

Different from the previous work~\cite{rossetti2017dynamic}, which had the purpose of achieving ``perfect” 3D  reconstruction  of  the  entire  WSI  stack,  we  consider  the registration  as  an  intermediate  tool  to  estimate  the  affinity between  glomerular  detection  results.  Therefore,  we  are  not aiming  to  align  all  sections  into  a  single  space,  but  only emphasize  the  registration  across  neighboring  sections(frames)  as  a  canonical  \ac{MOT}  setting. 

An important limitation of registration based tracking is the registration failure, which might break all global tracking identifications. When deploying \ac{Map3D} on the larger cohort, it is appealing that the algorithm itself could feedback the quality of registration across the series. Therefore, we propose the \ac{QaWS} registration method, to perform self-\ac{QA} to classify the quality of the registration on the entire series as ``good", ``acceptable", or ``bad". In this section, we focus on introducing the pair-wise registration in \ac{QaWS} registration, while the details of self-\ac{QA} method with cycle-consistent registration failure detection, are introduced in the next section, 

The pair-wise registration is employed to find the pixel-to-pixel correspondence between different pathological images. The correspondence is used to calculate the affinity score between detected objects using \ac{IoU}. Our registration consists of affine registration and non-rigid registration. For affine registration, scale-invariant feature transform (\ac{SIFT})~\cite{lowe2004distinctive}, speeded-up robust features (SURF)~\cite{bay2006surf}, and recent Graph Neural Network based SuperGlue (\ac{SG})~\cite{sarlin20superglue} are employed. For non-rigid registration, advanced normalization tools \ac{ANTs}~\cite{avants2008symmetric} is utilized due to the superior performance on high-resolution neuroimaging data.We define $t$ as the $t$-th  section (frame) in the entire series with length $T-2$, where $i$ is the index of pixel $x_i$ in the image $I$, with $N$ pixels.

\begin{equation}
M_{f_t} = \argminA_M \sum_{i=1}^N ||A(x_{i}^{t+1},M)-x_{i}^{t}||_{Aff}\label{pythagorean}
\end{equation}
\begin{equation}
\phi_{f_t} = \argminA_{\phi} \sum_{i=1}^N ||D[A(x_{i}^{t+1},M_{f_t}),\phi]-x_{i}^{t}||_{NR}\label{pythagorean1}
\end{equation}
\begin{equation}
T_{f_t} = (M_{f_t},\phi_{f_t})
\end{equation}


In Eq.(1) and (2), $A$ indicates the affine registration with initial parameters $M$, while $D$ represents the non-rigid \ac{ANTs} registration with optimized parameters $M_f$ and deformation field $\phi$. The $||.||_{Aff}$ and $||.||_{NR}$ in Eq.(1) indicate the different similarity metrics for affine registration and non-rigid registration, respectively. \ac{SIFT} and SURF used Euclidean distance between feature descriptors as the similarity metrics, while SuperGlue used Graph Neural Network layers to minimize the difference between each pair of keypoints. For \ac{ANTs} registration, both affine and non-rigid registration were performed with default similarity metrics of mutual information (MI) and cross correlation (CC), respectively. For the two-stage registration (e.g., \ac{SIFT}+\ac{ANTs} or \ac{SG}+\ac{ANTs}), the affine optimized parameters from SIFT or \ac{SG} will be employed to replace the affine registration stage in \ac{ANTs}({\color{red} \Fig~\ref{Fig.3.0}}). In Eq.(3), $T_f$ indicates the two-stage transformation from Eq.(1) and (2).

\subsection{Cycle-consistent Registration Failure Identification}
\begin{figure}
\centering 
\includegraphics[width=0.4\textwidth]{{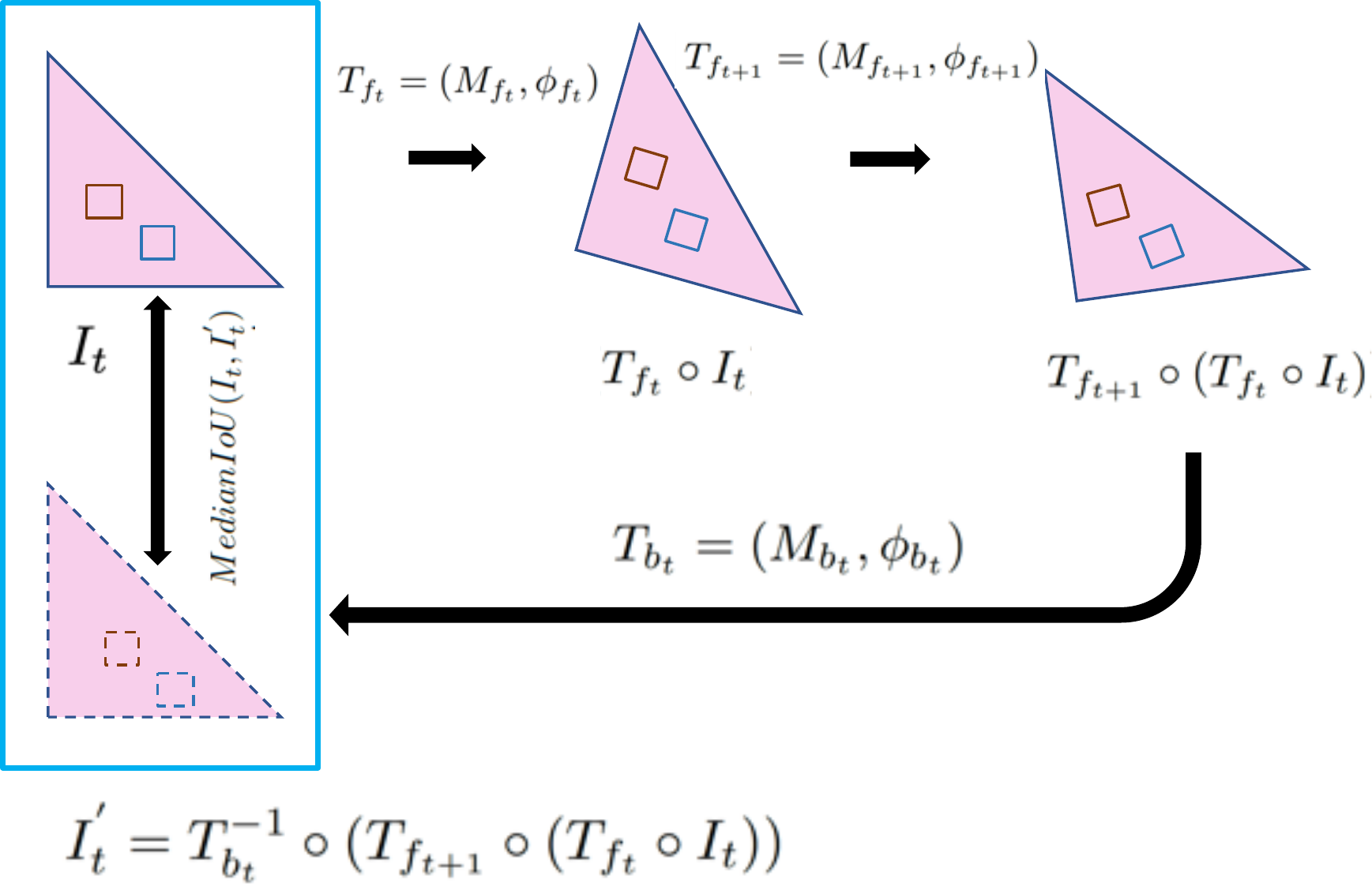}}
\caption{This figure presents the principle of cycle-consistent registration failure identification in \ac{QaWS} registration. If any pair-wise global registration failure happens in a cycle loop, the $I_t^{'}$ would not be well aligned with $I_t$, leading to small median \ac{IoU} score for bounding box pairs between $I_t^{'}$ and $I_t$.} 
\label{Fig.3.0} 
\end{figure}

To enable the quality-awareness for the entire series, we employ the additional registration pair between section $t+2$ and $t+1$ ({\color{red} \Fig~\ref{Fig.3.0}}). 

\begin{equation}
{M_f}_{t+1} = \argminA_M \sum_{i=1}^N ||A(x_{i}^{t+2},M)-x_{i}^{t+1}||_{Aff}\label{pythagorean2}
\end{equation}
\begin{equation}
{\phi_f}_{t+1} = \argminA_{\phi} \sum_{i=1}^N ||D[A(x_{i}^{t+2},M_{f_{t+1}}),\phi]-x_{i}^{t+1}||_{NR}\label{pythagorean3}
\end{equation}
\begin{equation}
{T_f}_{t+1} = (M_{f_{t+1}},\phi_{f_{t+1}})
\end{equation}

To form a cycle loop of the registration, we also perform an interleave registration from $t+2$ to $t$ in Eq.(4), (5), and (6). The $f_t$ and $f_{t+1}$ indicate the 1st and 2nd forward registration in the cycle-consistent registration failure detection, while the $b_t$ indicates the registration is performed for backward registration ({\color{red} \Fig~\ref{Fig.3.0}}).

\begin{equation}
M_{b_t} = \argminA_M \sum_{i=1}^N ||A(x_{i}^{t+2},M)-x_{i}^{t}||_{Aff}\label{pythagorean4}
\end{equation}
\begin{equation}
\phi_{b_t} = \argminA_{\phi} \sum_{i=1}^N ||D[A(x_{i}^{t+2},M_{b_t}),\phi]-x_{i}^{t}||_{NR}\label{pythagorean5}
\end{equation}
\begin{equation}
T_{b_t} = (M_{b_t},\phi_{b_t})
\end{equation}

With the affine registration parameters and deformation fields from the pair-wise and interleave registration, we will achieve the $I_t^{'}$ applying all affine and non-rigid deformation fields on the image $I_t$ as Eq.(10). The $'$ indicate applying a pair of affine and non-rigid registration, where three $\circ$ are from three independent registration procedures. Note that the inverse affine and deformation fields $T_{b_t}^{-1}$ are used to transfer the deformed image back to the original space in Eq.(10).


\begin{equation}
I_{t}^{'} = T_{b_t}^{-1} \circ ( T_{f_{t+1}} \circ (T_{f_t} \circ I_t))  
\end{equation}


To build our self-\ac{QA} algorithm for registration performance, we propose a new cycle-consistent registration failure identification. First, all automatic detection bounding boxes are transferred through the entire circle, all the auto detection bounding box from $I_t$ to $I_t^{'}$ . Then, we calculate the \ac{IoU} between each original bounding box and the deformed polygon (applying deformation fields on the bounding box) after the entire cycle loop. The median of \ac{IoU} score from all boxes is used as the cycle-consistency scores. In Eq.(8), we introduce a failed cycle-consistent $FC$ score to indicate the registration performance of each pair in our \ac{Map3D} algorithm. We set the cycle-consistency score threshold $Q$ = 0.1, where the \ac{IoU} above $Q$ means a successful registration pair ( as the threshold $Q$ of the fail cycle-consistent score $FC$=0). This threshold $Q$ = 0.1 was determined by a simulation that a set of bounding boxes were randomly shifted by 70 $\mu$m, to compute the \ac{IoU} with the original boxes. The random shifts simulated the levels of registration error. 70 $\mu$m~\cite{terasaki2020analysis} is empirically chosen as it is approximately the diameter of a mouse glomerulus. Therefore, the registration error larger than 70 $\mu$m is defined as a bad registration case.

\begin{equation}
FC_t = 
\begin{cases}
1, \quad MedianIoU(I_t,  I_{t}^{'}) < Q \\
0, \quad MedianIoU(I_t,  I_{t}^{'}) \ge  Q 
\end{cases}
\end{equation}


\subsection{Dual-path Association}
The \ac{DPA} algorithm is presented to associate all detected glomeruli across all sections in a 3D context using affinity measurements calculated from both adjacent and interleave registration. The \ac{IoU} is used to represent the affinity between detected bounding boxes and quadrilaterals. Two glomerular cross-sections with an \ac{IoU} score beyond a constant threshold $S$ are assigned with the same tracking number. The order of the \ac{IoU} association will follow the range from large to small, which eliminates the impact of false positve glomeruli with small \ac{IoU} scores. The choice of $S$ is provided in the ``Ablation Study" section. The \ac{DPA} algorithm is defined in Algorithm \ref{map3dtrack}.


\begin{algorithm}
\small
\caption{\ac{Map3D} Tracking Algorithm}
\label{map3dtrack}
\begin{algorithmic}[1]
\State \textbf{Stage 1:} Registration based affinity estimation.
\For{each section index $t$}
\State Register section $t$+1 to section $t$
\State Register section $t$+2 to section $t$
\State Cycle-consistent registration failure identification
\EndFor

\State
\State \textbf{Stage 2:} Dual-path association (\ac{DPA}) in 3D.
\State Assign the tracking number in the first section
\For{each section index $t$}
    \If{registration is successful between section $t$ and $t$+1}
        \State Assign the tracking numbers in section $t$+1 from section 
        \State $t$ using the largest \ac{IoU} affinity, for \ac{IoU}$>S$.
        \For{detected objects without tracking numbers in $t$+1}
        \State Assign the tracking numbers in section $t$+1 from 
        \State section $t$-1 using the largest IoU affinity, for IoU$>S$.
        \EndFor

        \State Assign new trackers for remaining objects in $t$+1
        \State $t$ = $t$+1
    \Else
        \State Assign the tracking numbers on section $t$+2 from section 
        \State $t$ using the largest IoU affinity, for IoU$>S$.
        \State Assign new trackers for remaining objects in $t$+2
        \State $t$ = $t$+2 
        
    \EndIf
    
\EndFor

\end{algorithmic}
\end{algorithm}

\section{Experimental Design}
\subsection{Data}

14 mouse kidney sections (3D volumes) have been digitized from two previous studies\footnote{\label{note1}Kidney Int. 2017 Dec;92(6):1395-1403.}$^{,}$\footnote{\label{note2}Kidney Int. 2020 Nov 1;S0085-2538(20)31240-0.}. DTR mice (C57 bl/6 background), which express $\gamma$-glutamyl transferase 1 diphtheria toxin receptor (Ggt1 DTR) on proximal tubular epithelial cells, were induced tubular injury by human DT (Sigma, 100 ng/kg body weight i.p.) injection. Patchy tubular injury and interstitial fibrosis were induced by folic acid (FA, 40 $\mu$g/g, i.p.; Sigma Aldrich, St Louis, MO) injection. All animal procedures were approved by the Institutional Animal Care and Use Committee at Vanderbilt University.

Each mouse kidney was prepared through staining with hematoxylin for nuclei and lectin for proximal tubule detection. Seven to seventeen 8 $\mu$m sections were cut through each mouse kidney. WSI was acquired for all sections at 20$\times$ magnification (0.5 $\mu$m pixel resolution) by using Leica SCN400 Slide Scanner. Images were saved as .scn files. All used mice are presented in {\color{green} \Tab~\ref{table4}}.

\begin{figure*}
\begin{center}
\centering 
\includegraphics[width=1\textwidth]{{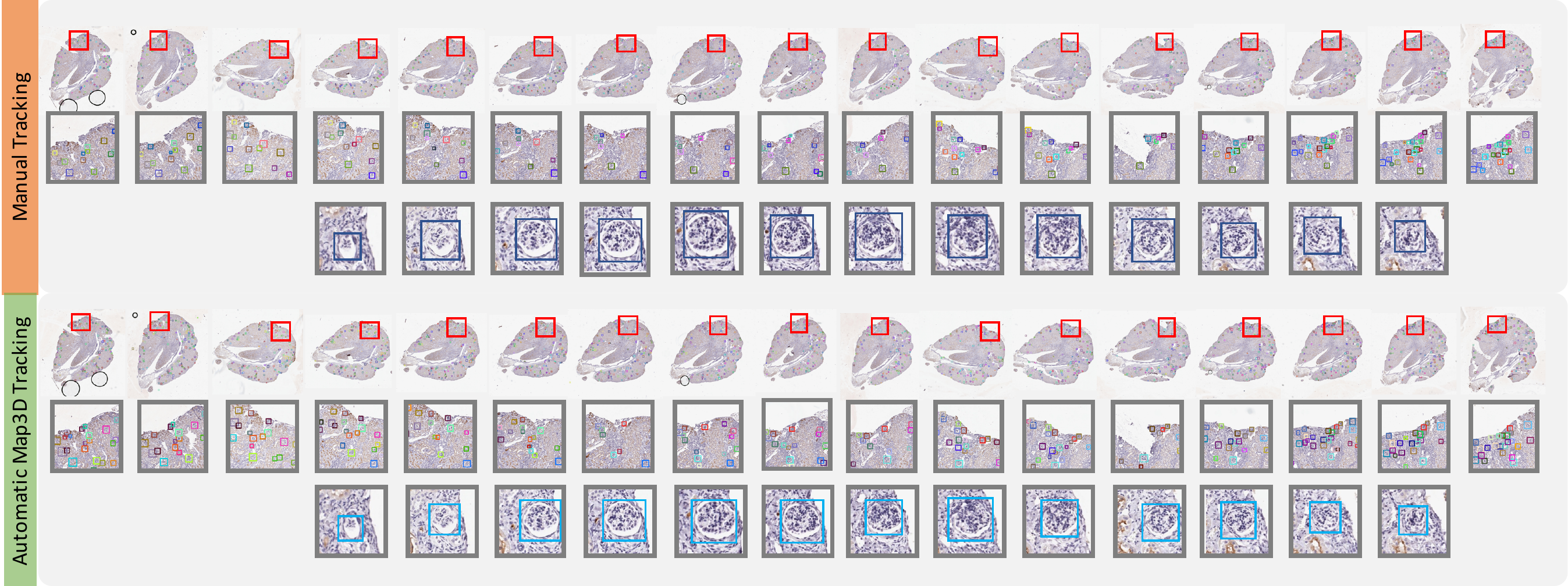}}
\end{center}
\caption{This figure shows the tracking results on the testing images. The 17 sections are obtained from the same mouse kidney.The upper panel shows the manual annotation, which takes 30 hours of human effort, while the lower panel shows the automatic tracking results using our proposed \ac{Map3D} method with no human effort. In each panel, the first row is the WSI. The second row is the enlarged region within the red boxes in WSI, where different colors indicate different tracking identifications. The third row is all cross-sections from one tracked 3D glomerulus, where the same color is assigned with the same tracking identification.} 

\label{Fig.3} 
\end{figure*}

\subsection{Experimental Design}
To evaluate the performance of object tracking, we manually annotated a single mouse kidney with the largest number of sections (17 sections), and placed manual bounding boxes and tracking numbers across all 17 sections. The entire manual process of annotating and \ac{QA} one single kidney took 30 hours of human labor. This sample included 297 glomeruli and 1605 2D detection results. The manual annotation was saved as sections indices, detection coordinates, and tracking numbers. 

To choose the optimal hyper parameters for registration and association, we manually annotated and tracked another mouse kidney as the validation data. For our validation purposes, we only annotated three  adjacent representative sections, with missing tissue, to save on manual effort. The validation images consisted of 66 glomeruli from 172 2D detection results. This validation kidney was used to determine the optimal threshold $S$ of the \ac{IoU} when associating detection results (see ``Ablation Study"). 

Aside from the testing and validation kidneys, we chose another 10 kidneys (8 training and 2 validation) for fine-tuning the CircleNet detection method. Since the data was only used for detection, our pathologist only annotated one 2D whole section from each kidney. As the results, 927 and 125 glomerular detection results were manually annotated as training and validation data. As all glomeruli were annotated on the testing kidney, the testing data with 1605 glomerular detection results were used to evaluate final detection performance.

To perform the ablation study of comparing different registration methods, our pathologist manually traced one 3D glomerulus from each of 12 kidneys, to form 178 sections for registration with 95 adjacent pairs and 83 interleave pairs of sections for registration error calculation. 

\begin{table}
\caption{Data Information.}
\centering
\begin{tabular}{ccccc}
\hline
 ID&Name&Sections&Study&utilization\\
\hline
  1&13-260&12&DT&Registration Error\\
  2&13-261&10&DT&Registration Error\\
  3&13-262&8&DT&Registration Error\\
  4&13-263&10&DT&Registration Error\\
  5&13-265&9&DT&Registration Error\\
  6&13-268&8&DT&Registration Error\\
  7&13-270&9&DT&Registration Error\\
  8&13-274&7&DT&Registration Error\\
  9&13-315&10&FA&Registration Error\\
  10&14-7X&8&FA&Registration Error\\
  11&14-81&8&FA&Registration Error\\
  12&189553&8&DT&Registration Error\\
  13&189550&3&DT&Validation\\
  14&189552&17&DT&Testing\\
  
\hline
\end{tabular}
\label{table4} 
\end{table}

\subsection{Evaluation Metrics}

\textbf{MOT.} We use the standard \ac{MOT} metrics for Multi-Target Tracking from \ac{MOT}-Challenge 2015~\cite{MOTChallenge2015} to verify our tracking results. All manual annotation and automatic tracking results are saved in the \ac{MOT}-Challenge 2015 format to be compatible with the official evaluation code. IDF1, IDP, and IDR (the larger, the better) are the ratio of correctly identified detections over the number of ground-truth and computed detections according to the F1, recall, and precision scores. MOTA, MOTP, and MOTAL (the larger, the better) are the multiple object tracking accuracies, which combine false positives, missed targets, and identity switches metrics.

\textbf{Registration.} In the ablation study, we also evaluate the registration performance by using the absolute distance between landmarks. The registration error is calculated between the center points of the corresponding manual glomerular detection results, using the absolution distance. 
 

\begin{table*}
\caption{\ac{MOT} performance on manual annotation results.}
\centering
\begin{tabular}{p{2cm}p{0.25cm}p{0.25cm}p{0.25cm}p{0.25cm}p{0.25cm}p{0.25cm}p{0.25cm}p{0.25cm}p{0.25cm}p{0.25cm}p{0.25cm}p{0.25cm}p{0.25cm}p{0.25cm}p{0.5cm}p{0.5cm}p{0.5cm}}
 \hline
 Method& IDF1 &IDP&IDR&Rcll&Prcn&FAR&GT&MT&PT&ML&FP&FN&IDs&FM&M1&M2&M3\\
 \hline
  SURF~\cite{bay2006surf}&76.2&76.2&76.2&100&100&0.00&297&297&0&0&0&0&198&16&87.0&90.3&99.8\\
  SIFT~\cite{SIFT}&85.6&85.6&85.6&100&100&0.00&297&297&0&0&0&0&101&16&93.4&90.3&99.9\\
  SuperGlue(SG)~\cite{sarlin20superglue}&93.1&93.1&93.1&100&100&0.00&297&297&0&0&0&0&45&16&97.0&90.3&99.9\\ 
  ANTs ~\cite{Avants2011ARE}&75.6&75.6&75.6&100&100&0.00&297&297&0&0&0&0&177&16&88.3&90.3&99.9\\
  Map3D(S+A)&97.4&97.4&97.4&100&100&0.00&297&297&0&0&0&0&7&16&\textbf{99.5}&90.3&99.9\\
  Map3D(SG+A)&97.4&97.4&97.4&100&100&0.00&297&297&0&0&0&0&7&16&\textbf{99.5}&90.3&99.9\\
  Map3D(S+A+D)&\textbf{98.9}&\textbf{98.9}&\textbf{98.9}&100&100&0.00&297&297&0&0&0&0&7&16&\textbf{99.5}&90.3&99.9\\
  Map3D(SG+A+D)&\textbf{98.9}&\textbf{98.9}&\textbf{98.9}&100&100&0.00&297&297&0&0&0&0&7&16&\textbf{99.5}&90.3&99.9\\ 
 \hline
  \text{*S is SIFT, A is ANTs, D is DPA}\\
 \text{*M1 is MOTA, M2 is MOTP, M3 is MOTAL}
\end{tabular}
\label{table1}
\end{table*}

\begin{table*}
\caption{\ac{MOT} performance on automatic detection results.}
\centering
\begin{tabular}{p{2cm}p{0.25cm}p{0.25cm}p{0.25cm}p{0.25cm}p{0.25cm}p{0.25cm}p{0.25cm}p{0.25cm}p{0.25cm}p{0.25cm}p{0.25cm}p{0.25cm}p{0.25cm}p{0.25cm}p{0.5cm}p{0.5cm}p{0.5cm}}
\hline
 Method& IDF1 &IDP&IDR&Rcll&Prcn&FAR&GT&MT&PT&ML&FP&FN&IDs&FM&M1&M2&M3\\
 \hline
  SURF~\cite{bay2006surf}&60.7&52.8&71.5&90.1&66.5&43.1&297&234&51&12&689&150&165&48&33.9&70.6&44.6\\
  SIFT~\cite{SIFT}&67.4&58.6&79.4&90.1&66.5&43.1&297&234&51&12&689&150&75&48&39.8&70.6&44.6\\
  SuperGlue(SG)~\cite{sarlin20superglue}&71.5&62.1&84.2&90.1&66.5&43.1&297&234&51&12&689&150&35&48&42.5&70.6&\textbf{44.7}\\ 
  ANTs~\cite{Avants2011ARE}&59.3&51.6&69.8&90.1&66.5&43.1&297&234&51&12&689&150&164&48&34.0&70.6&44.6\\
  Map3D(S+A)&74.7&64.9&88.0&90.1&66.5&43.1&297&234&51&12&689&150&6&48&44.4&70.6&\textbf{44.7}\\
  Map3D(SG+A)&74.5&64.8&87.8&90.1&66.5&43.1&297&234&51&12&689&150&7&48&44.3&70.6&\textbf{44.7}\\
  Map3D(S+A+D)&\textbf{75.3}&\textbf{65.4}&\textbf{88.6}&90.1&66.5&43.1&297&234&51&12&689&150&3&48&\textbf{44.6}&70.6&\textbf{44.7}\\
  Map3D(SG+A+D)&75.1&65.3&88.4&90.1&66.5&43.1&297&234&51&12&689&150&4&48&44.5&70.6&\textbf{44.7}\\
 \hline
 \text{*S is SIFT, A is ANTs, D is DPA}\\
 \text{*M1 is MOTA, M2 is MOTP, M3 is MOTAL}
\end{tabular}
\label{table2}
\end{table*}

\section{Results}
\subsection{MOT}
We performed a standard \ac{MOT} evaluation on the testing data ({\color{green} \Tab~\ref{table1} and \ref{table2}}) following the definitions of the metrics in ~\cite{MOTChallenge2015}. To disentangle the effects of detection and tracking components, we evaluated the final results using both (1) manual annotation, and (2) automatic detection. The large-scale results are shown in {\color{red} \Fig~\ref{Fig.3}}.

\textbf{\ac{MOT} with Manual Annotation.} In this scenario, the manual annotation ({\color{green} \Tab~\ref{table1}}) was used to show the \ac{MOT} performance when the detection is ``perfect". The proposed \ac{Map3D} with \ac{DPA} achieved the best performance. 

\textbf{Detection Results.} The automatic detection results are also presented in {\color{green} \Tab~\ref{table2}}. The CircleNet achieved 90.1 \% recall (Rcll) and 66.5 \% precision (Prcn). As the detection is not the focus of this paper, the comprehensive analyses of the detection can be found in~\cite{yang2020circle}.

\begin{figure}
\includegraphics[width=0.47\textwidth]{{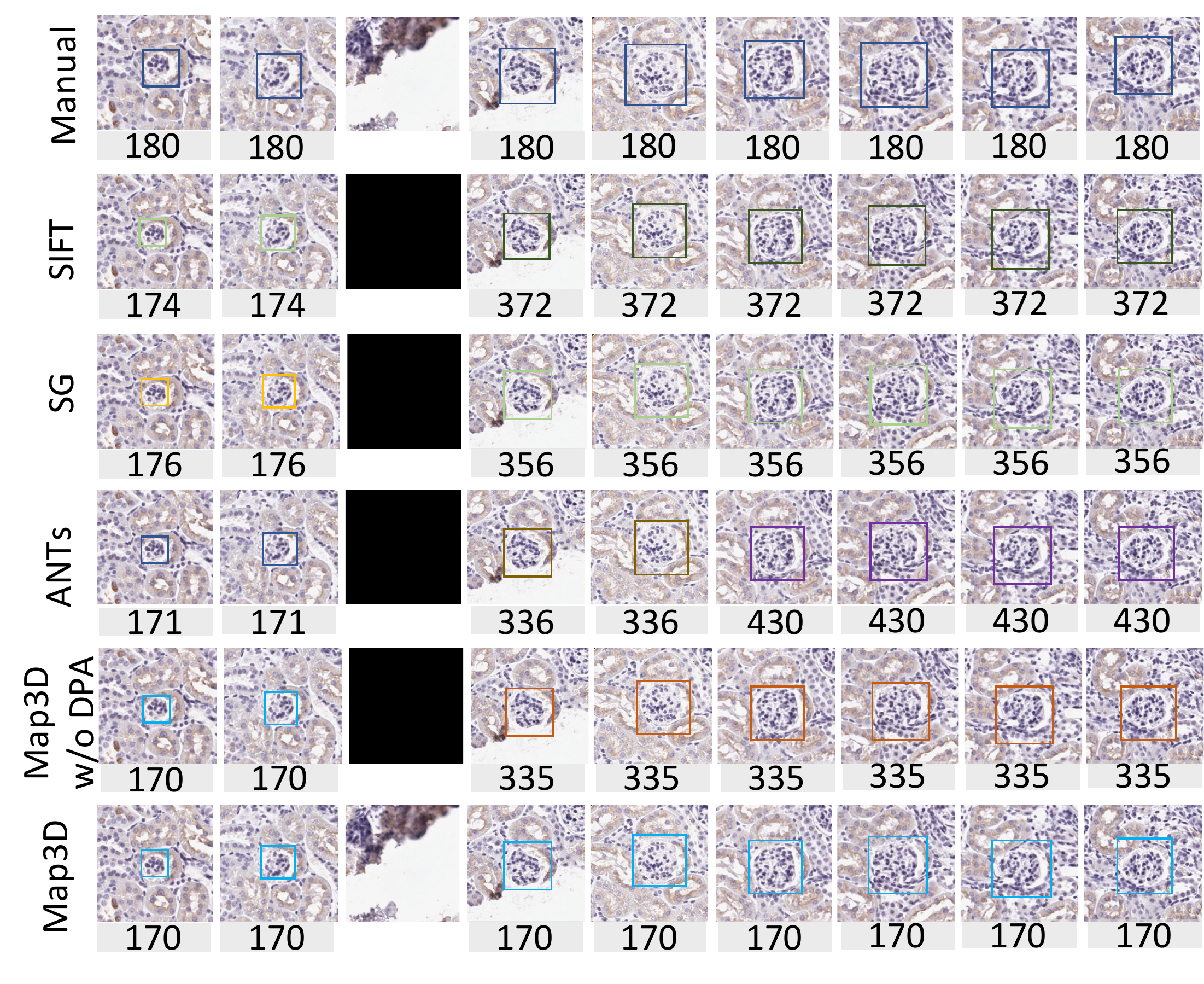}}
\caption{This figure demonstrates the tracking results with missing tissue and incomplete glomeruli. The numbers below images are the tracking identifications. The proposed \ac{Map3D} method with \ac{DPA} was able to achieve consistent tracking results by skipping over the missing tissues.} 
\label{Fig.4} 
\end{figure}

\textbf{\ac{MOT} with Automatic Detection.} In this experiment, the automatic detection results from  CircleNet was used as the detection results ({\color{green} \Tab~\ref{table2}}). The results showed that the proposed \ac{Map3D} with \ac{DPA} also achieved the best performance, compared with baseline methods. In {\color{red} \Fig~\ref{Fig.4}}, the proposed \ac{Map3D} is able to achieve more consistent tracking results by skipping over the missing tissues, artifacts, and false negative results in automatic detection.

\textbf{\ac{MOT} between different metrics.} In our experiment, the performance of \ac{MOT} (IDF1, IDP, IDR, and MOTA \cite{ristani2016performance}) is largely affected by the association because the same detection results are used. For the same reason, the MOTP and MOTAL \cite{bagdanov2012posterity} are insensitive across different methods. When different detection results are used (({\color{green} \Tab~\ref{table1}}) vs. ({\color{green} \Tab~\ref{table2}})) the MOTP and MOTAL are informative.

\subsection{Ablation Studies}
Using the validation set, the tracking results  with different \ac{IoU} thresholds $S$ are presented in {\color{green} \Tab~\ref{table3}} using the same \ac{Map3D} tracking methods with manual annotation. The tracking results with threshold $S$ = 0.1 achieved the best performance with 98.8 in IDF1, compared with ground truth tracking. Since the evaluation was performed on manual annotation, the detection related metrics are not provided in {\color{green} \Tab~\ref{table3}}.

\begin{table}
\caption{Tracking performance for different \ac{IoU} Threshold on validation images using \ac{Map3D}.}
\centering
\begin{tabular}{cccccccc}
\hline
 IoU& IDF1 &IDP&IDR&IDs&MOTA&MOTP&MOTAL\\
\hline
  0.1& \textbf{98.3}&\textbf{98.3}&\textbf{98.3}&\textbf{2}&\textbf{98.8}&89.4&\textbf{99.7}\\
  0.2& 97.7&97.7&97.7&3&98.3&89.4&99.6\\
  0.3& 95.9&95.9&95.9&6&96.5&89.4&99.5\\
  0.4& 89.0&89.0&89.0&18&89.5&89.4&99.3\\
  0.5& 80.2&80.2&80.2&36&79.1&89.4&99.1\\
\hline
\end{tabular}
\label{table3} 
\end{table}

\begin{figure}
\centering 
\includegraphics[width=0.45\textwidth]{{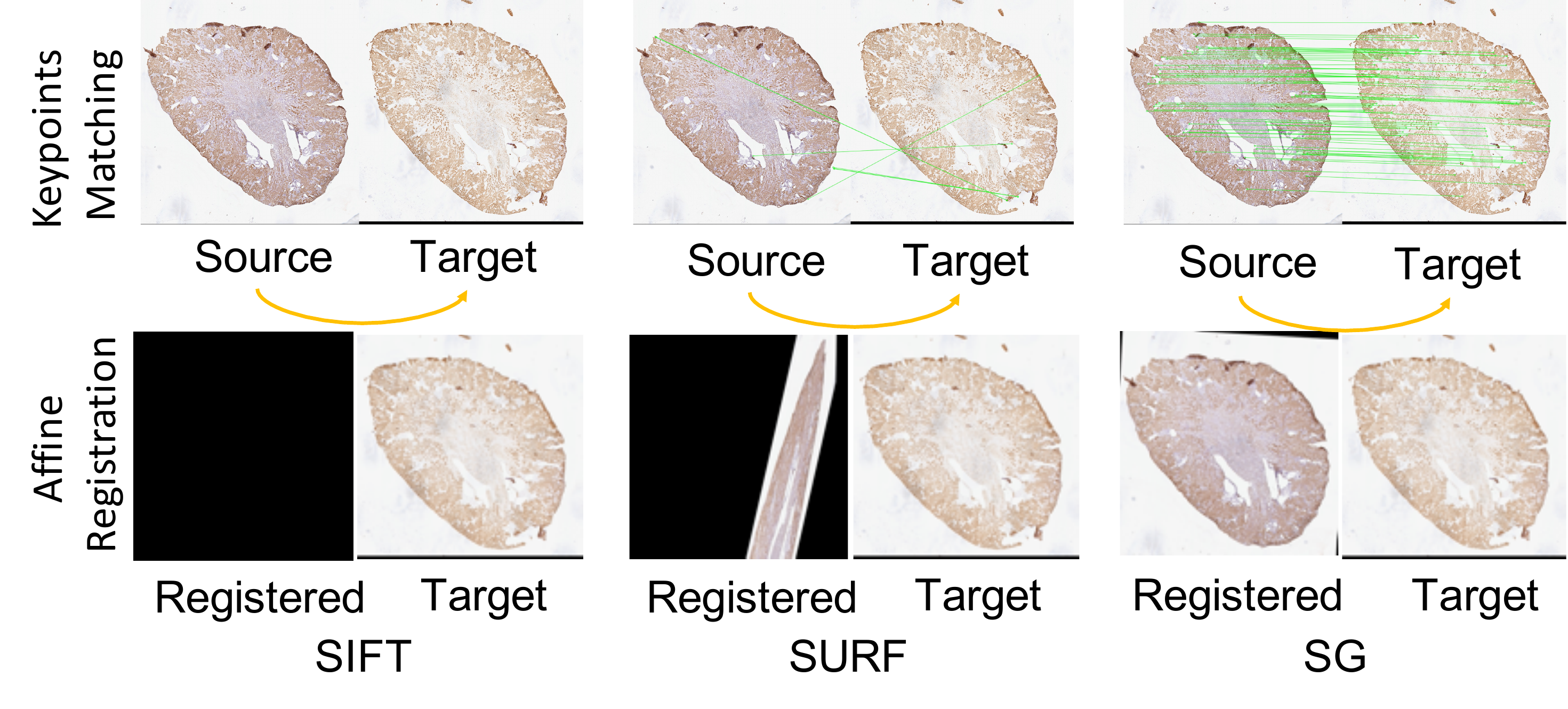}}
\caption{The affine registration performance is evaluated and presented with three algorithm(i.e., \ac{SIFT}, SURF, and \ac{SG}) on one pair of images with large contrast variations.} 
\label{Fig.8} 
\end{figure}

\begin{figure}
\centering 
\includegraphics[width=0.4\textwidth]{{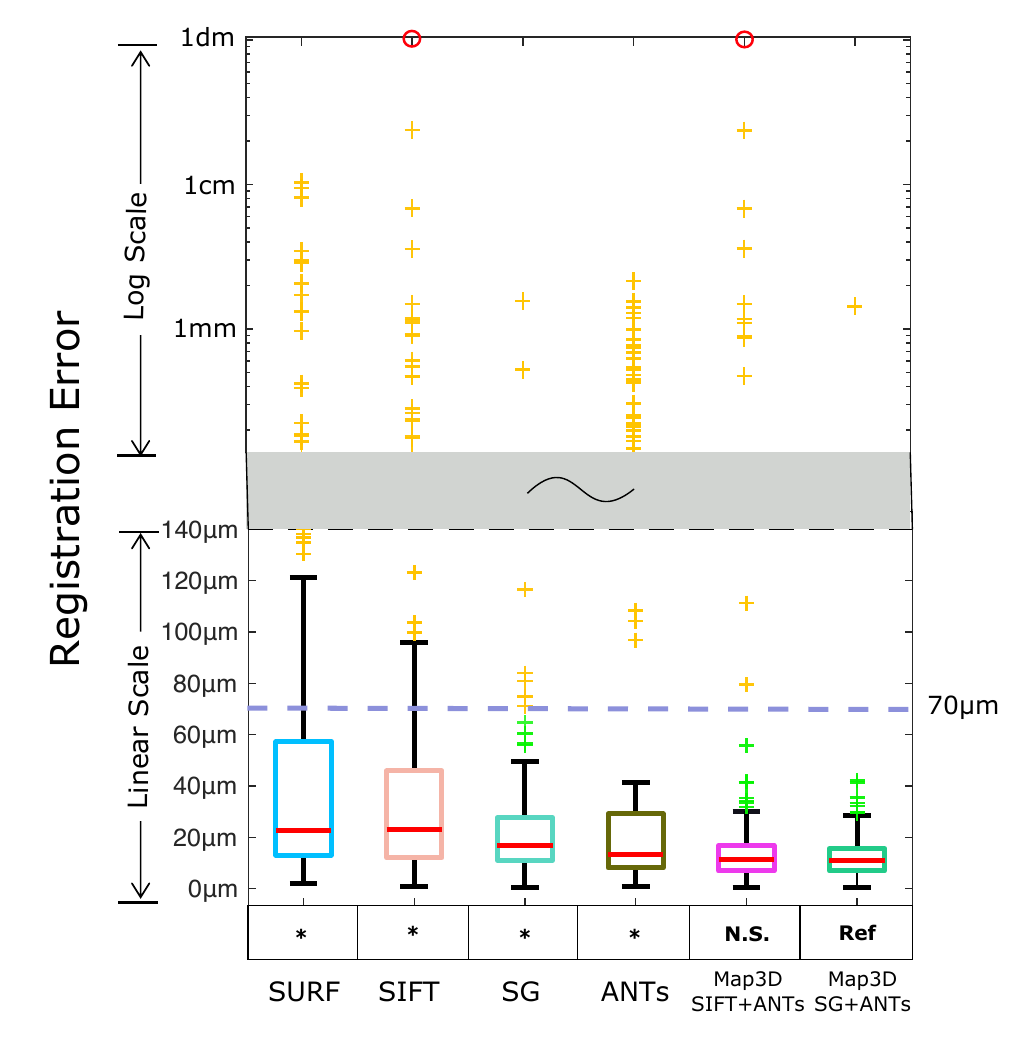}}
\caption{The figure shows the pair-wise registration errors between 178 pair of consequtive slides. 70 $\mu$m is approximately the diameter of a mouse glomerulus. Therefore, the ones with registration error larger than 70 $\mu$m (dashed line) are defined as ``bad" registration cases. The yellow crosses are the ones with the registration error larger than 70 $\mu$m, while the red circles are the failed registrations without affine outcomes. The Wilcoxon signed-rank test is performed with \ac{Map3D} (\ac{SG}+\ac{ANTs}) as the reference (``Ref”) method, to compare with other methods. ``*" represents the significant (p $<$ 0.05) differences, while ``N.S." means the difference is not significant.} 
\label{Fig.6} 
\end{figure}

 {\color{red} \Fig~\ref{Fig.8}} indicates an example of using \ac{SIFT}, SURF and SuperGlue during affine registration. The green lines link the matched keypoints, which shows that the deep learning based solution (SuperGlue) is more robust for intensity and contrast variations. {\color{red} \Fig~\ref{Fig.6}} also shows the performance of pairwise registration and \ac{QaWS} registration on 178 pairs of sections from 12 kidneys. As shown in the figure, the two-stage registration used in the \ac{Map3D} with Super Glue achieved the lowest median registration error. 
 
\begin{figure}
\centering 
\includegraphics[width=0.3\textwidth]{{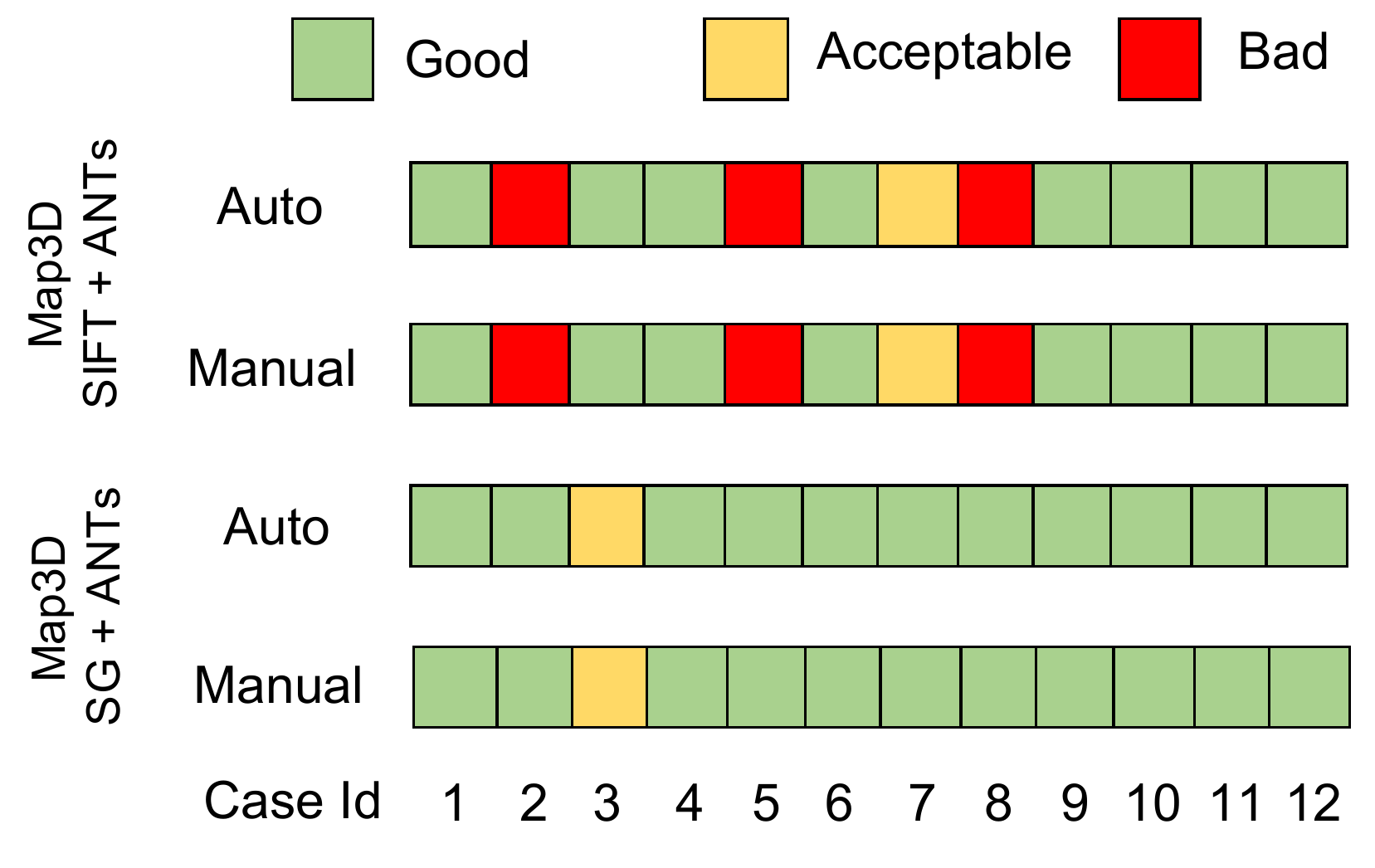}}
\caption{The figure shows the automatic whole series self-\ac{QA} results with different registration methods. ``Good" means no global registration failure is detected. ``Acceptable" is defined as no more than two consecutive pairs of sections have $FC = 1$. ``Bad" corresponds to more than one consecutive pair of section that has $FC = 1$ } 
\label{Fig.7} 
\end{figure}

The automatic \ac{QA} from \ac{QaWS} registration is presented in  {\color{red} \Fig~\ref{Fig.7}}. The \ac{QA} is evaluated at the kidney level, which covers all sections in each kidney. The mean and standard deviation of the longer boundaries (width or height) of all automatic bounding boxes in self-QA are 87 $\mu$m and 17 $\mu$m, respectively. The mean size is comparable with the size of objects (the average diameter of glomeruli is 70 $\mu$m). For \ac{QA} results, we define ``good” as no global registration failure is detected, which means the registrations are good for all pairs across all sections. ``Acceptable” is defined as no more than two consecutive pairs of sections have $FC$=1, which means such registration failure is acceptable for our \ac{Map3D} algorithm as the tracking can not be built by skipping the bad registration pairs ({\color{red} \Fig~\ref{Fig.4}}). Such results are acceptable for downstream tasks, such as calculating 3D volume, percentage of sclerosis etc. If more than one consecutive pair of section has $FC$=1, the results are ``bad” for \ac{Map3D} as the consistent tracking can not be built. For ``bad" cases, the pathologists might need to manually align the problematic pairs, indicated by $FC$ scores.

{\color{green} \Tab~\ref{table5}} compares the results of using box and circle representation of the same detection results from CircleNet. Box here means to obtain a minimal square around the circle results. The \ac{IoU} and Circle\ac{IoU} threshold are all set to 0.1. The bounding box representation achieves better results in IDF1, IDP, and IDR, while the circle representation has slightly better results in MOTA. The bounding box representation is used in this study to be consistent with other detection and \ac{MOT} methods.

\begin{table*}
\caption{\ac{MOT} performance with circle representation.}
\centering
\begin{tabular}{p{2.5cm}p{0.25cm}p{0.25cm}p{0.25cm}p{0.25cm}p{0.25cm}p{0.25cm}p{0.25cm}p{0.25cm}p{0.25cm}p{0.25cm}p{0.25cm}p{0.25cm}p{0.25cm}p{0.25cm}p{0.5cm}p{0.5cm}p{0.5cm}}

 \hline
 Method&IDF1&IDP&IDR&Rcll&Prcn&FAR&GT&MT&PT&ML&FP&FN&IDs&FM&M1&M2&M3\\
 \hline
  Map3D in MA w R&\textbf{98.9}&\textbf{98.9}&\textbf{98.9}&100&100&0.00&297&297&0&0&0&0&7&16&99.5&90.3&99.9\\
  Map3D in MA w C&98.0&98.0&98.0&100&100&0.00&297&297&0&0&0&0&6&16&\textbf{99.6}&90.3&99.9\\ 
  Map3D in AD w R&\textbf{75.1}&\textbf{65.3}&\textbf{88.4}&90.1&66.5&43.1&297&234&51&12&689&150&4&48&44.5&70.6&44.7\\
  Map3D in AD w C &73.9&64.2&87.0&90.1&66.5&43.1&297&234&51&12&689&150&3&48&\textbf{44.6}&70.6&44.7\\
 \hline
 \text{*Map3D is Map3D(\ac{SG} + ANTs + DPA, MA means manual annotation, AD is automatic detection}
 \text{*R is rectangle bounding box, C is circle representation}
 \text{*M1 is MOTA, M2 is MOTP, M3 is MOTAL}
\end{tabular}
\label{table5}
\end{table*}

\section{Discussion}

In this study, we offer new capabilities of investigating glomeruli in 3D, by deriving the 3D glomerular quantification \ac{Map3D} as a \ac{MOT} problem using routine serial sectioning and assessment from WSI. As a holistic solution, deep learning based detection and registration-based tracking enables the previously infeasible large-scale glomerular associations across 3D sections. 

In {\color{green} \Tab~\ref{table1} and \ref{table2}}, our \ac{Map3D} registration based association achieved better \ac{MOT} performances compared with the classical registration methods. Our two-stage registration yielded superior registration performance with smaller registration errors, which are mostly less than 70$\mu$m (the average diameter of a glomerulus) in {\color{red} \Fig~\ref{Fig.6}}. Our Dual-path Association tackled missing areas of tissues and artifacts across sections ({\color{red} \Fig~\ref{Fig.4}}), whereas the proposed self-\ac{QA} automatically detected the registration failures ({\color{red} \Fig~\ref{Fig.7}}).

One promising application of the proposed \ac{Map3D} algorithms is to assist the identification of atubular glomerulus in a mouse kidney, by tracking individual glomeruli in 3D. The computer assisted quantification would reduce the human efforts of screening serial sections from WSIs.\footnote{https://www.asn-online.org/education/kidneyweek/2020/program-abstract.aspx?controlId=3448300}

Another promising future direction is to apply the proposed method to other types of histological tissues (e.g., cancer pathological tissues) or different imaging modalities (e.g., fluorescent microscopy).

There are several limitations and potential future improvements in the current version of \ac{Map3D}. First, one major limitation is the computational cost for non-rigid registration. Currently, more than five minutes are required to perform a pair-wise registration, which would take hours to conduct all necessary dual-path registrations for extended serial sections. The deep learning based non-rigid registration methods can be introduced to the \ac{Map3D} framework, which can be even further combined with detection as a holistic algorithm~\cite{liu2020jssr}.

Second, since the pairwise affinity estimation is a standard setting in \ac{MOT}, we use a chain-like strategy to associate nearby sections for tracking objects. However, this strategy is not able to localize the cases that are of bad quality or are missing tissue happens in more than two consecutive sections. In the future, a star-like strategy \cite{rossetti2017dynamic} which associates all sections would be a promising solution to further improve the performance. Meanwhile, the affinity estimation is performed between detected boxes and transformed boxes. To further improve the performance, the non-local patch search~\cite{huo2017simultaneous} widely used in the Multi-atlas Segmentation (MAS) could be included to further enhance the tracking performance. 

Third, the comparison between manual annotation and automatic detection indicates that the detection performance is  another critical factor in determining the performance of overall tracking. Because of the variety of similar shapes between vessels and glomerulus on kidney images, the automatic detection has low precision. It might be solved by increasing the size of the dataset from other histological slices, which also help us enhance the applicability further.

In addition, the detection results, if precise enough, may be used as landmarks and descriptors to further improve the performance of registration and association in a joint optimization manner. This would also be used as a spatial prior since the glomeruli are naturally 3D objects. However, the optimization of glomeruli detection is beyond the scope of the present study. In {\color{green} \Tab~\ref{table2}}, we used the same automatic detection results (even if not perfect), to compare different tracking methods as a fair comparison. With the same detection setting, the proposed method achieved consistently better tracking results.

\section{Conclusion}
In this paper, we propose the \ac{Map3D} method, to approach the large-scale glomerular identification and association in 3D serial sections from a \ac{MOT} perspective. The proposed \ac{Map3D} consists of a glomerular detection, quality-aware \ac{QaWS} registration, and a dual-path 3D association method \ac{DPA}. \ac{Map3D} achieves superior tracking performance compared with baseline methods in the large-scale glomerular association, tackling potential issues such as the missing area of tissues in sections and artifacts. 

\section*{Acknowledgment}
This work was supported in part by NIH NIDDK DK56942(ABF).

\ifCLASSOPTIONcaptionsoff
  \newpage
\fi



\bibliographystyle{IEEEtran}
\bibliography{main}
%







\end{document}